 \documentclass[pmlr,twocolumn]{jmlr} 

\usepackage{booktabs}
\usepackage[load-configurations=version-1]{siunitx} 

\theorembodyfont{\upshape}
\theoremheaderfont{\scshape}
\theorempostheader{:}
\theoremsep{\newline}

\jmlrvolume{ML4H Extended Abstract Arxiv Index}
\jmlryear{2020}
\jmlrsubmitted{2020}
\jmlrpublished{}
\jmlrworkshop{Machine Learning for Health (ML4H) 2020}

\title[Evaluation of a machine learning model in pharmacy]{Comparison of pharmacist evaluation of medication orders with predictions of a machine learning model}

  \author{%
   \Name{Sophie-Camille Hogue} \Email{sophie-camille.hogue.hsj@ssss.gouv.qc.ca}\\
   \Name{Flora Chen} \Email{flora.chen.hsj@ssss.gouv.qc.ca}\\
   \Name{Geneviève Brassard} \Email{genevieve.brassard.hsj@ssss.gouv.qc.ca}\\
   \Name{Denis Lebel} \Email{denis.lebel.hsj@ssss.gouv.qc.ca}\\
   \Name{Jean-François Bussières} \Email{jean-francois.bussieres.hsj@ssss.gouv.qc.ca}\\
   \Name{Maxime Thibault} \Email{maxime.thibault.hsj@ssss.gouv.qc.ca} \\
  \addr CHU Sainte-Justine, Montréal, Canada 
\AND
\Name{Audrey Durand} \Email{audrey.durand@ift.ulaval.ca}\\
\addr Université Laval, Québec, Canada
}

\begin{document}

\maketitle

\begin{abstract}
The objective of this work was to assess the clinical performance of an unsupervised machine learning model aimed at identifying unusual medication orders and pharmacological profiles. We conducted a prospective study between April 2020 and August 2020 where 25 clinical pharmacists dichotomously (typical or atypical) rated 12,471 medication orders and 1,356 pharmacological profiles. Based on AUPR, performance was poor for orders, but satisfactory for profiles. Pharmacists considered the model a useful screening tool.
\end{abstract}

\section{Introduction}

  Pharmacists ensure the adequate use of medication. Under the most frequent pharmacy practice model in North American hospitals, pharmacists review the lists of active medications (pharmacological profiles) for inpatients under their care. This meticulous process aims to identify medications that require intervention or further verification. However, most medication orders do not contain drug related problems~\citep{Mahoney2007-ja}. Publications from more than 10 years ago illustrate the potential of technology to help pharmacists streamline repetitive processes such as medication order review~\citep{Flynn2009-xa,Tribble2009-hx,Poikonen2009-lx}. Recent works discussed theoretical concepts of machine learning (ML) and their potential uses in pharmacy, but also pointed out the lack of published studies demonstrating applications~\citep{Nelson2020-pk,Flynn2019-sh}.

  An approach based on ML could augment and support the capabilites of pharmacists by focusing their attention on orders more likely to be problematic, as it has been shown that errors made by pharmacists increase with the number of prescriptions to be verified per shift~\citep{Gorbach2015-hv}.

  In this prospective study, we evaluated the performance of a ML model for identifying medication orders and pharmacological profiles that are atypical, defined as deviating from usual prescription patterns. The study took place in a tertiary-care mother-and-child academic hospital between April 2020 and August 2020. 
  
\section{Methods}
  
  \subsection{Dataset and models}

    A dataset consisting of 2,846,502 medication orders from 2005 to 2018 was extracted from the pharmacy database and preprocessed to reconstruct 1,063,173 pharmacological profiles. We were unable to find previous work that performed anomaly detection in medication lists. Therefore, we used this dataset to implement unsupervised anomaly detection techniques. Technical details of training and evaluation are available in \appendixref{apd:model_dev}. The code for all the models  with all final parameters is available online\footnote{\url{https://github.com/pharmai-lab/PharmAI_2}}. The protocol and access to the data were approved by the local research ethics committee.

    We constructed a statistical baseline on the hypothesis that profiles considered as typical by pharmacists would correspond to those that they had seen frequently in the past. As a second baseline, we considered four ML techniques for anomaly detection: robust covariance, one-class support vector machines, isolation forests and local outlier factor based on $k$-nearest neighbors. Only the isolation forest model produced clinically realistic results as determined by a pharmacist, and therefore other models were discarded. We then constructed a neural network autoencoder, on the hypothesis that a properly constructed model should be able to easily reconstruct common profiles but not atypical ones. Because such a model outputs a prediction for each dimension of the output vector, we hypothesized that we could use these predictions as anomaly scores for individual drugs within profiles. We finally used this autoencoder in an adversarial training model adapted from GANomaly~\citep{Akay2018GANomalySA}, an autoencoder-based generative adversarial network. We hypothesized that adversarial training would improve predictions by inciting the reconstructions to resemble real profiles as much as possible.

    Since GANomaly showed the best performance during development (see~\appendixref{apd:model_dev}), this model was selected for display to pharmacists. Only individual medication predictions from GANomaly were evaluated. For profiles, predictions from all models were compared to a pharmacist opinion. Before data collection and monthly, the models were retrained with the 10 years of most recent data in order to minimize drift.
    
    \begin{figure}[htbp]
        \floatconts
         {fig:pr_curves}
         {\caption{Precision-recall curves for profiles}}
         {\includegraphics[width=0.48\textwidth]{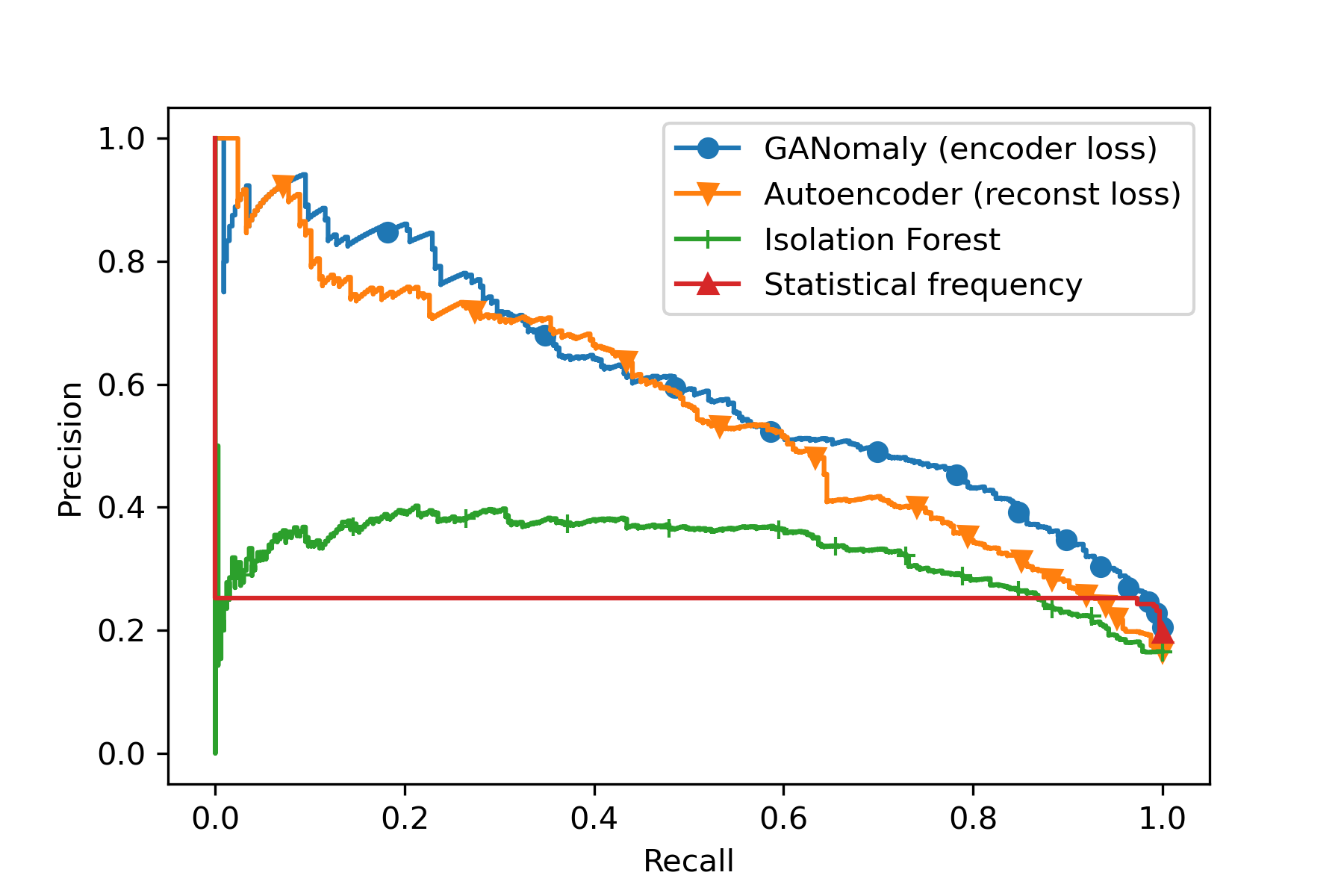}}
    \end{figure}

  \begin{table*}
  \floatconts{tab:characteristics}%
    {\caption{Characteristics of pharmacists, ratings, and performance by department}}%
    {%
    \setlength{\tabcolsep}{3pt}
    \begin{tabular}{lllllll}
      \toprule
      Department &  Pharmacists, & Experience & Orders, & Profiles, & F1 & F1\\ 
      & n & (years), & n (\%) & n (\%) & orders & profiles \\
      && mean $\pm$ std &&&& \\
      \midrule    
      Overall &  25   & 10.7 $\pm$ 7.4   & 12,471 (100) & 1,356 (100) & 0.30 & 0.59 \\
      \midrule
      Ob-gyn &  4   & 15.5 $\pm$ 6.8   & 7,119 (57) & 536 (40) & 0.25 & 0.55 \\
      General ped &  4   & 13.5 $\pm$ 9.0   & 2,036 (16) & 368 (27) & 0.28 & 0.66 \\
      Surgery &  2   & 2.0 $\pm$ 1.4   & 1,113 (9) & 107 (8) & 0.45 & 0.42 \\
      Oncology &  6   & 10.0 $\pm$ 6.9   & 731 (6) & 32 (2) & 0.32 & 0.65 \\
      Specialized ped &  3   & 6.0 $\pm$ 8.7   & 615 (5) & 67 (5) & 0.39 & 0.69 \\
      NICU &  5   & 12.0 $\pm$ 8.5   & 410 (3) & 99 (7) & 0.13 & 0.35 \\
      Nursery &  N/A *   &    & 267 (2) & 128 (9) &  &  \\
      PICU &  1   & 13   & 180 (2) & 19 (1) & 0.33 & 0.72 \\
      \bottomrule
      \multicolumn{7}{l}{*\footnotesize{Not applicable, were rated by pharmacists from ob-gyn}}
    \end{tabular}
    }
  \end{table*}
  
  \subsection{Data collection}

    Daily, profiles for newly admitted patients were included. A patient could only be evaluated once, to minimize the risk of including profiles that the pharmacist had previously evaluated. Pharmacists rated every medication order (typical or atypical) before observing the predictions. An atypical prescription was defined as one that did not correspond to usual prescribing patterns, according to the pharmacist’s expertise. A profile was considered atypical if at least one medication order within the profile was rated as atypical. Predictions were then displayed to the pharmacists and they indicated whether they agreed or disagreed with the profile prediction. This agreement was used to evaluate performance by profile after the display of predictions. We explored the perceptions of AI in healthcare and usefulness of this model during focus groups with pharmacists.  
  
\section{Results}\label{sec:results}

    12,624 medication orders and 2,114 profiles were displayed to 25 pharmacists from seven departments. 153 (1.2\%) medication orders and 758 (36\%) profiles were excluded. 699 (33\%) of these profiles were excluded because they were used to calculate the classification threshold based on encoder loss during the first two months (see \appendixref{apd:model_dev} for details). A total of 12,471 orders and 1,356 profiles were included in the analysis (\tableref{tab:characteristics}).

    Medication order predictions generated by the GANomaly model displayed poor performance, with a precision of 35\%, recall of 26\%, specificity of 97\%, and negative predictive value of 96\% (\tableref{tab:confusion}). F1-score was only 0.30 (F1-scores by department are shown in \tableref{tab:characteristics}) and area under precision-recall (AUPR) was only 0.25.
    
    Profile predictions generated by the GANomaly model, when compared with pharmacist ratings before seeing predictions, achieved satisfactory performance with a precision of 49\%, recall of 75\%, specificity of 82\%, and negative predictive value of 93\% (\tableref{tab:confusion}). F1-score was 0.59. (F1 scores by department are shown in \tableref{tab:characteristics}). All F1-scores for profiles showed improvement compared to those for orders, except in one department. Predictions for profiles by GANomaly obtained the best performance compared with baselines, with an AUPR of 0.60, while the autoencoder had an AUPR of 0.56 and the isolation forest 0.33 (\figureref{fig:pr_curves}). When pharmacists were asked if they agreed with predictions, precision was 66\%, recall 87\%, specificity 87\%, and negative predictive value 96\% (\tableref{tab:confusion}). F1-score rose to 0.75, showing that pharmacists agreed well with predictions.

  \begin{table*}
    \floatconts{tab:confusion}%
    {\caption{Confusion matrices of pharmacist ratings compared to predictions}}{
    \begin{tabular}{llllllll}
      \toprule
      &&\multicolumn{2}{c}{Individual orders} & \multicolumn{2}{c}{Profiles before} & \multicolumn{2}{c}{Profiles after} \\
      &&&& \multicolumn{2}{c}{seeing predictions} & \multicolumn{2}{c}{seeing predictions}\\
      \cmidrule(r){3-4}\cmidrule(r){5-6}\cmidrule{7-8}
      && Atypical & Typical & Atypical & Typical & Atypical & Typical \\
      \midrule    
      Model &  Atypical   & 166 & 304 & 195 & 201 & 263 & 133 \\
      & Typical & 465 & 11,536 & 66 & 894 & 38 & 922 \\ 
      \bottomrule
    \end{tabular}
    }
  \end{table*}
    Over the 15 pharmacists who participated in the focus groups, 9 preferred predictions by medication order without being aware that performance was better with profiles. The model was useful to prioritize patients and to identify dosage form errors.
  
\section{Discussion}\label{sec:discussion}

  The GANomaly model proved to be the best approach for pharmacological profile predictions among those evaluated. However, most pharmacists preferred predictions for individual orders. Conceptually, presenting pharmacists with a prediction for each order should be better because it identifies clearly which prescription is atypical, unlike profile predictions which only inform the pharmacist that something is atypical within the profile. Although focus groups indicated a lack of trust in order predictions by pharmacists, they were satisfied to use them as a safeguard to ensure that they did not miss unusual orders. This leads us to believe that even moderately improving the quality of these predictions in future work could be beneficial.

  Two methods~\citep{Woods2014-vt,Santos2019-ox} have been described in previous work to identify atypical medication orders with ML. These methods used the generic name, the route of administration, the dose and the frequency of administration to determine if a dose, route and frequency for a drug were statistically common or not. Our approach is different as it analyzes medications in relation to one another. It could be combined with these techniques to offer more complete predictions.

  Being a single-centre study, the model was trained only with data from our centre and generalization to another institution is not possible. We believe that reproducing these results across institutions would be an issue since the trained model finely models practice patterns. However, it may be possible to train institution-specific GANomaly models and to compare results across institutions.

  We do not believe that the GANomaly model could be used as a standalone decision support tool because ordering patterns include legitimate atypical orders as well as suboptimal, but common, practices. However, it could be combined with classical rule-based approaches to detect such issues independently of practice patterns.

  Pharmacists from obstetrics-gynecology rated almost half of the orders and  profiles in this study, over-representing this population. Also, some pharmacists indicated that they may have rated orders based on how clinically appropriate they found them rather than how much they corresponded to known patterns. This may have lowered performance metrics.

\section{Conclusion}

  This study evaluated the clinical performance of a ML model capable of identifying atypical medication orders and pharmacological profiles. It showed better performance for the identification of atypical profiles than orders. Improving these predictions should be prioritized in future research to maximize clinical impact.

\bibliography{ml4h_2020}

\appendix

\section{Model development}\label{apd:model_dev}

We partitioned the model development dataset into a training and validation set composed of all orders between 2005 and 2017 inclusively (n = 989,766 profiles containing [mean $\pm$ std] 9.4 $\pm$ 6.9 drugs, 4,329 unique drugs), and a test set of all orders from 2018 (n = 73,407 profiles of 10.11 $\pm$ 7.6 drugs, 2,280 unique drugs). Reconstructed profiles were composed of database medication identifiers (a unique string representing the drug and its dosage form) but did not include the dose, route or frequency of administration, as other approaches have been developed for these elements (see \sectionref{sec:discussion}).

\subsection{Statistical frequency }

We counted the occurrences of each unique pharmacological profile in the training set. When using this model on new data, each profile was attributed a score corresponding to the inverse (such that rare profiles would have a higher score) of the frequency of that exact profile in the training set.

\subsection{Anomaly detection techniques}

For these models, profiles were represented as count vectors of drugs within profiles, on which latent semantic indexing was performed. Each technique was tested on the resulting latent representation.

Based on a preliminary, in-house study, we observed that pharmacists rated between 10 and 20\% of pharmacological profiles as atypical overall. As such, we considered a fixed proportion of anomalies of 20\% in the training/validation dataset during training.  We performed experiments to explore the effect of varying the training data volume (1 to 4 years), the number of components of the truncated singular-value decomposition part of the latent semantic indexing (8 to 1024 components, explaining from ~20\% to ~90\% of data variance), and the model type. We examined performance by looking at the proportion of outliers on the validation set in selected patient populations, namely neonatal intensive care (NICU), obstetrics and gynecology (ob/gyn), oncology and general pediatrics. We considered that an acceptable model would show a lower proportion of outliers in NICU and ob/gyn patients than in oncology and general pediatrics, because these populations have more protocolized orders and generally follow more predictable medication ordering patterns. 

\begin{figure*}
    \floatconts{fig:enc_thr}{\caption{Encoder loss thresholds}}
    {\includegraphics[width=1\textwidth]{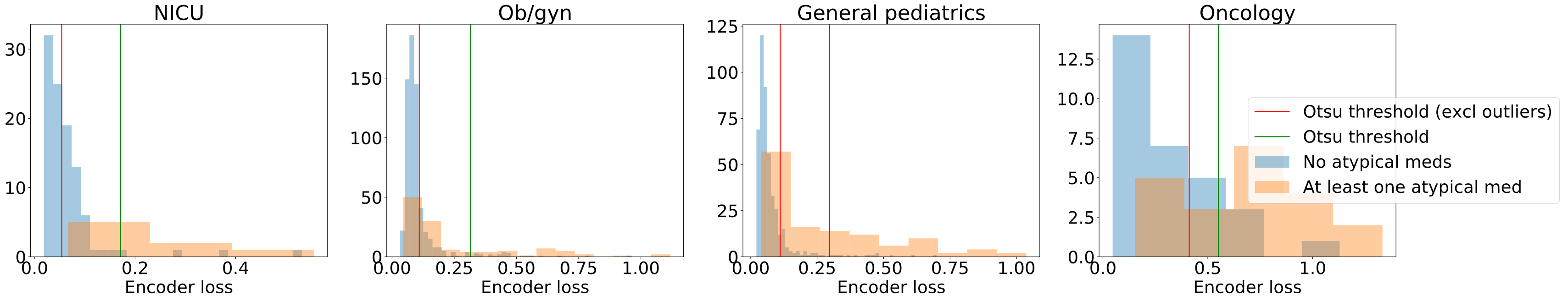}}
\end{figure*}

We performed this experiment with 3-fold cross-validation where years 2015, 2016 and 2017 were used as the three validation sets and the preceding 1 to 4 years were used as the training sets. We ensured that pharmacological profiles for each hospitalization did not leak between sets by including data for each entire hospitalization in the year in which it began, even if it ended in the following year or later.

We observed that 512 components for the isolation forest method appeared to offer the most clinically valid results with a (mean $\pm$ std) overall anomaly ratio of 15.7 $\pm$ 0.6 \%, 10.0 $\pm$ 1.7\% in ob/gyn, 5.3 $\pm$ 0.6\% in NICU, 16.3 $\pm$ 0.8\% in general pediatrics and 20.6 $\pm$ 1.3\% in oncology. Other methods did not provide clinically realistic results. This final model was trained on the year 2017 and evaluated on the test set. The overall ratio of atypical pharmacological profiles was 18.4\%, with 7.6\% in ob/gyn, 9.7\% in NICU, 20.2\% in general pediatrics and 28.1\% in oncology.

\subsection{Neural network autoencoder}

We represented profiles as multi-hot vectors and used this representation directly as input to the model. For all experiments for this model, we used the same 3-fold cross-validation schema as previously described.

We first established a structure able to reconstruct profiles with between 10 and 20\% of individual orders flagged as atypical on validation data. The autoencoder structure we selected was composed of three fully-connected layers of respectively 256, 64 and 256 nodes with ReLU activation and a dropout rate of 0.1 after each 256-node layer. The model was trained with the Adam optimizer with a 10\textsuperscript{-3} learning rate and a batch size of 256. During experiments on the training and validation partition, early stopping occurred when the validation loss decreased less than 10\textsuperscript{-4} over 5 epochs. For the final model trained with no validation, based on results of experiments with early stopping, training occurred for 11 epochs.

We tested increasing the training data volume up to 10 years and observed increased performance. The validation reconstruction accuracy on 10 years of training data was (mean $\pm$ std) 88.8 $\pm$ 1.1 \% and the validation rate of atypical orders was 9.6 $\pm$ 0.6 \%. On the test set, the reconstruction accuracy of the model was 88.0 \% and the rate of atypical orders was 9.9 \%.

\subsection{GANomaly-based model}

In this adversarial model, anomalies are detected by computing the loss, called encoder loss, between two encoders of the same structure, respectively taking as input the original data and its reconstruction. This value would, in our case, be used to detect profile anomalies.

We explored the hyperparameters of the model until we obtained validation metrics that resembled those of the original autoencoder. We modified the original loss functions by replacing contextual loss with binary cross-entropy instead of L1 loss, we replaced the encoder loss with a weighted L2 + L1 loss instead of L2 loss, and we kept the adversarial loss as L2 loss. The feature extractor structure was one layer of 128 nodes followed by one layer of 64 nodes. We used Adam optimizers with a 10\textsuperscript{-3} learning rate for the encoder-decoder-encoder structure and 10\textsuperscript{-6} learning rate for the feature extractor. We used SELU activation with a 0.1 dropout ratio after 256-nodes layers in the encoder-decoder-encoder structure and ReLU activation with batch normalization and no dropout in the feature extractor. These modifications helped stabilize training and avoid mode collapse, which occurred quickly otherwise.

The loss weights were 100 for contextual loss, 2 for adversarial loss and 1 for encoder loss. The encoder loss weights were 0.8 for L1 loss and 0.2 for L2 loss. For all experiments with this model, we used the same 3-fold cross-validation schema as previously described. During experiments on the training and validation partition, early stopping occurred as previously. For the final model trained with no validation, training occurred for 21 epochs. 

This model resulted in a validation reconstruction accuracy of (mean $\pm$ std) 88.8 $\pm$ 1.1 \% and a ratio of atypical orders of 9.7 $\pm$ 0.8 \%. On the test set, the reconstruction accuracy was 87.4\% and the ratio of atypical orders 11.1\%.

In order to display predictions to pharmacists, it was necessary to dichotomize the encoder loss value in an unsupervised manner. We selected Otsu’s method for thresholding. Because this method needed to be applied on the encoder loss distribution from unseen data, we used the first two months of encoder loss values generated during the study to calculate the thresholds and excluded these profiles from the main analysis. We calculated a threshold for each population of patients, since the distribution varied by population. We observed that outlier values seemed to disproportionately influence the threshold when examining values separated by pharmacist opinion about the presence of at least one atypical medication in the profile. For this reason, we excluded encoder loss values above the 90th percentile for each department from threshold calculation. The results of this experiment for 4 selected departments are presented in \figureref{fig:enc_thr} (histograms are separated by pharmacist opinion using results from the study for clarity, but the thresholding method does not use pharmacist opinion).
\end{document}